\documentclass[a4paper,twoside]{article}

\usepackage{epsfig}
\usepackage{subcaption}
\usepackage{calc}
\usepackage{amssymb}
\usepackage{amstext}
\usepackage{amsmath}
\usepackage{amsthm}
\usepackage{multicol}
\usepackage{pslatex}
\usepackage{apalike}
\usepackage{algorithm2e}
\usepackage[bottom]{footmisc}

\usepackage{booktabs} 
\usepackage{graphicx}
\usepackage{hyperref}

\usepackage{SCITEPRESS}     

\begin{document}

\title{Oil Spill Segmentation using Deep Encoder-Decoder Models}

\author{\authorname{Abhishek Ramanathapura Satyanarayana\orcidAuthor{0009-0003-1248-0988} and Maruf A. Dhali\orcidAuthor{0000-0002-7548-3858}}
\affiliation{Department of Artificial Intelligence, Bernoulli Institute, University of Groningen, 9747 AG Groningen, The Netherlands}
\email{\{a.ramanathapura.satyanarayana@student.rug.nl, m.a.dhali@rug.nl}
}


\abstract{Crude oil is an integral component of the world economy and transportation sectors. With the growing demand for crude oil due to its widespread applications, accidental oil spills are unfortunate yet unavoidable. Even though oil spills are difficult to clean up, the first and foremost challenge is to detect them. In this research, the authors test the feasibility of deep encoder-decoder models that can be trained effectively to detect oil spills remotely. The work examines and compares the results from several segmentation models on high dimensional satellite Synthetic Aperture Radar (SAR) image data to pave the way for further in-depth research. Multiple combinations of models are used to run the experiments. The best-performing model is the one with the ResNet-50 encoder and DeepLabV3+ decoder. It achieves a mean Intersection over Union (IoU) of $64.868\%$ and an improved class IoU of $61.549\%$ for the ``oil spill" class when compared with the previous benchmark model, which achieved a mean IoU of $65.05\%$ and a class IoU of $53.38\%$ for the ``oil spill" class.}

\onecolumn \maketitle \normalsize \setcounter{footnote}{0} \vfill

\section{\uppercase{Introduction}}
\label{sec:introduction}
Seas, oceans, and coastal regions represent vital components of the environment, marine ecosystems, and human activities. The aquatic ecosystem is a crucial source of economic stability and livelihoods for a significant portion of the global population. However, numerous human activities pose substantial threats to marine ecosystems, with oil spills being a notable one. These spills can arise from various sources, including the transportation of crude oil. Crude oil is integral to many industrial and manufacturing processes, finding application in sectors such as gasoline, diesel, jet fuel, lubricants, textiles, paint, fertilizers, pesticides, and pharmaceuticals, serving as a critical driver of industrial development and expansion. However, the primary method of transporting crude oil globally is through shipping tankers, which inevitably leads to accidental oil spills.

Apart from accidental oil spills, other causes may include incidents from offshore platforms, drilling rigs and wells, natural disasters, deliberate releases, and technical failures. These events can lead to catastrophic environmental impacts, adverse human health effects, and significant socio-economic consequences. Additionally, oil spills can severely affect wildlife, including birds and marine mammals, disrupting ecosystems and leading to long-term ecological damage \cite{bird_die_1}. Besides affecting marine ecosystems, it also affects the air quality \cite{oil_spill_air_quality}. Oil spills have faced significant public and media backlash due to their harmful effects, prompting political and governmental institutions to prevent future occurrences \cite{steps_to_prevent_oil_spills}. 

Oil spills may take a long time to clean up, and the duration can vary significantly depending on several factors, including the spill's size, the type of oil, environmental conditions, and the effectiveness of the response efforts \cite{shigenaka2009hindsight}. It is essential to automatically and efficiently detect oil spills to enable prompt action for containment and cleanup. One effective approach is to combine remote sensing with artificial intelligence and supervised machine learning models specifically trained to identify spills. Remote sensing can be accomplished through satellite imagery for this purpose. Recently, there has been a growing interest in using deep Convolutional Neural Networks (CNNs) to process this image data. The introduction of the AlexNet model has demonstrated significant performance improvements over traditional feature engineering techniques, particularly in the ImageNet object recognition competition. \cite{imagenet_old,imagenet_new,alex_net}. 

Earlier works have utilized CNN models to detect oil spills using Synthetic Aperture Radar (SAR) satellite imagery. Marios Krestenitis et al. applied some modifications and trained some of the CNN models to detect oil spills in their research \cite{OilSpillDataset}. They divided the original high-dimensional images into smaller patches and trained various models on these. This process may consume a significant amount of memory, necessitating extensive computing resources and resulting in increased energy usage. As technology advances, developing and training models directly on higher-dimensional images is essential to achieve better performance. This can help reduce the memory required for processing by a certain amount, reducing the energy required for computation. In this research, an attempt is made to train various CNN models on relatively higher dimensional images and study the effects on the overall performance of the models.

\section{\uppercase{Related works}\label{sec:related_work}}
With the emergence of Artificial Neural Networks (ANNs), Yann LeCun et al. proposed LeNet-5, the first Convolutional Neural Network (CNN), that was applied to the image digit recognition task \cite{lenet_old,lenet_new}. Later, a popular dataset --- the MNIST dataset became a benchmark for the digit recognition task in images \cite{mnist}. There was a significant gap in the application of CNNs to various Computer Vision and Image Analysis tasks for multiple reasons, with a lack of computation power and a lack of large-scale labeled datasets being a few among them. Recently, ImageNet has been hosting an object recognition challenge on a large-scale dataset \cite{imagenet_old,imagenet_new}. This dataset consists of 1.2 million images belonging to 1000 classes. In 2012, a CNN model named AlexNet won the ImageNet object recognition challenge, outperforming all the other participants of that year by a large margin \cite{alex_net}. The other participants used non-CNN methods, i.e., the traditional handcrafted features combined with different machine learning techniques.

One key challenge in computer vision is semantic segmentation, where the model must learn to classify each pixel in an image into a specific category. In 2014, Jonathan Long et al. proposed a Fully Convolutional Network (FCN), which was the first CNN with only convolutional and transposed convolutional layers and without any fully connected (or dense) layers \cite{fcn}. With the advent of FCNs, researchers have proposed and developed several other state-of-the-art models for the semantic segmentation task. Among them, the popular ones include --- UNet, LinkNet, PSPNet, DeepLabV3+ \cite{unet,link_net,psp_net,deeplabv3+}. These models have one thing in common --- i.e., they are all variations of encoder-decoder architectures. Most of these models have been benchmarked on large datasets such as COCO and Cityscapes \cite{coco,cityscapes}.

Cityscapes, a labeled dataset used as a benchmark for semantic segmentation tasks, contains $5000$ high quality labeled images and $20000$ weakly labeled images. Marios Krestenitis et al. benchmarked some of these models on the Oil Spill Detection Dataset, a dataset developed by compiling the images extracted from the satellite Synthetic Aperture Radar (SAR) data \cite{OilSpillDataset}. This dataset contains a meager $1002$ training image, which is relatively much lower than the Cityscapes dataset. In their research, Marios Krestenitis et al. trained the models by dividing the original images of size $1250 \times 650$ into multiple patches. They used the image patches as input for the models. They used different image patch sizes for various models. Although different encoders were used in the original implementations of these models, Marios Krestenitis et al. modified the models to use ResNet-101 encoder for most of the decoders, the exception being MobileNetV2 encoder with DeepLabV3+ decoder in their research \cite{res_net,OilSpillDataset,deeplabv3+,mobile_net_v2}. Their study found that the MobileNetV2 encoder coupled with the DeepLabV3+ decoder scored the highest mean Intersection over Union (m-IoU) of $65.06\%$. They also found that this model scored the second highest class IoU of $53.38\%$ for the oil spill class.


\section{\uppercase{Methodology}\label{sec:methodology}}
\subsection{Dataset}\label{sec:dataset}
\begin{figure}[t]
    \centering
    \includegraphics[width=0.9\linewidth]{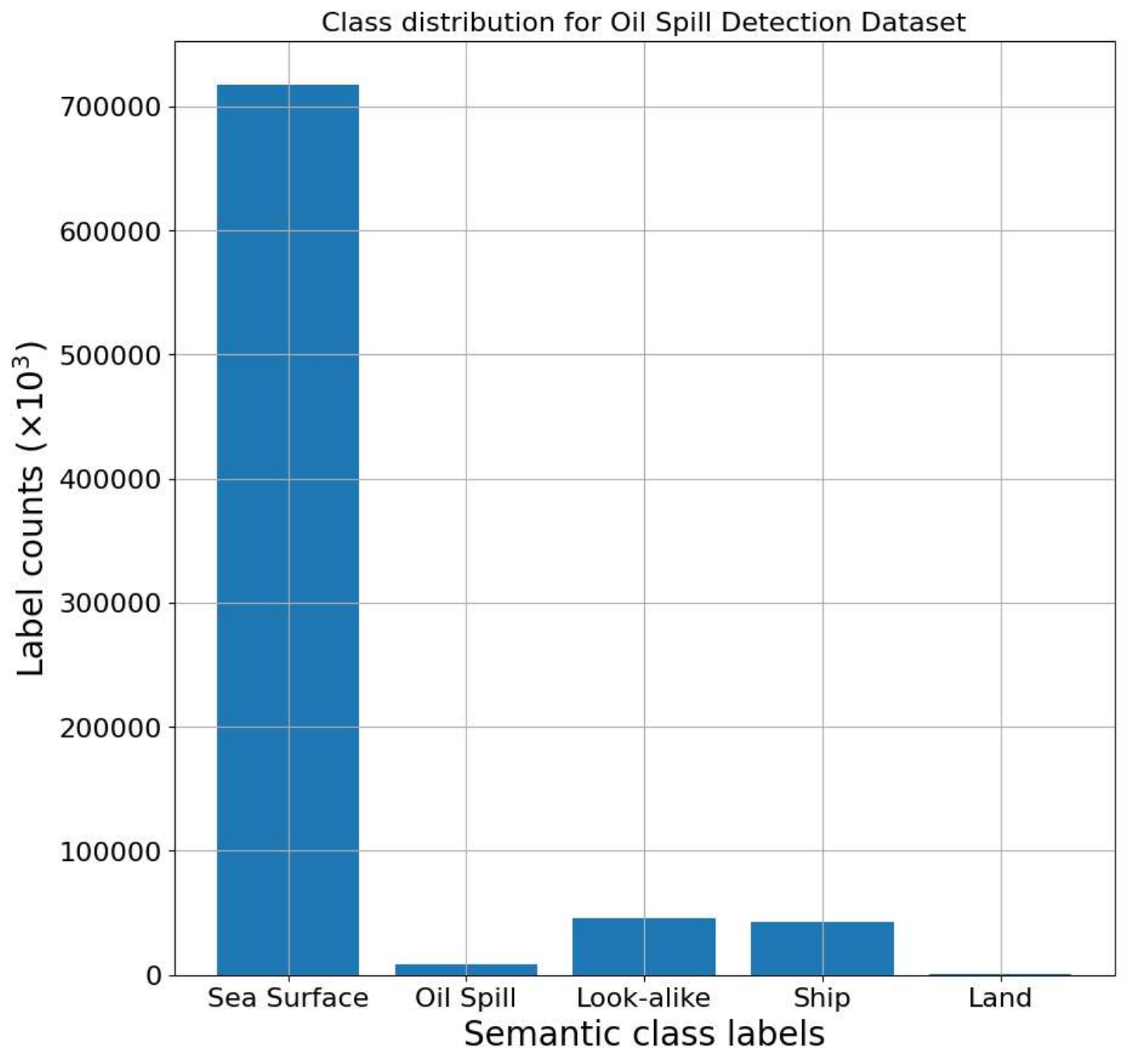}
    \caption{Plot showing the distribution of the semantic classes in the Oil Spill Detection Dataset.}
    \label{fig:class_distribution}
\end{figure}

Marios Krestenitis et al. developed the Oil Spill Detection Dataset, used in this research \cite{OilSpillDataset}. The dataset consists of images extracted from satellite Synthetic Aperture Radar (SAR) data depicting oil spills and other relevant semantic classes and their corresponding ground truth masks and labels \cite{OilSpillDataset}. It has $1002$ images in the training set and $110$ in the test set, with corresponding labels. There are $5$ semantic classes representing the following types --- sea surface, oil spill, oil spill look-alike, ship, and land. Figure \ref{fig:class_distribution} shows the class distribution in the training set. It can be observed that the dataset is highly imbalanced concerning the various semantic classes. For example, the number of pixels belonging to the ``Sea Surface" class outnumbers the other labeled semantic classes in the dataset. The class of interest in this particular research is ``Oil Spill", whose occurrence in the dataset is shallow. In this research, we attempt to train models optimized for detecting this class.

The images in the dataset are $1250 \times 650$ in $W \times H$, where $W$ and $H$ represent the width and height of the image, respectively. In the work of this article, the training dataset ($1002$ images), provided by default, was further split randomly into $95\%$ ($951$ images) and $5\%$ ($51$ images) for --- training and validation sets, respectively. This was done to perform a 5-fold cross-validation with randomized validation sets.

\subsection{Image Preprocessing \& Data Augmentation}\label{sec:image_preprocessing}
\begin{figure}[t]
    \centering
    \begin{subfigure}[b]{0.25\textwidth}
        \centering
        \includegraphics[width=\linewidth]{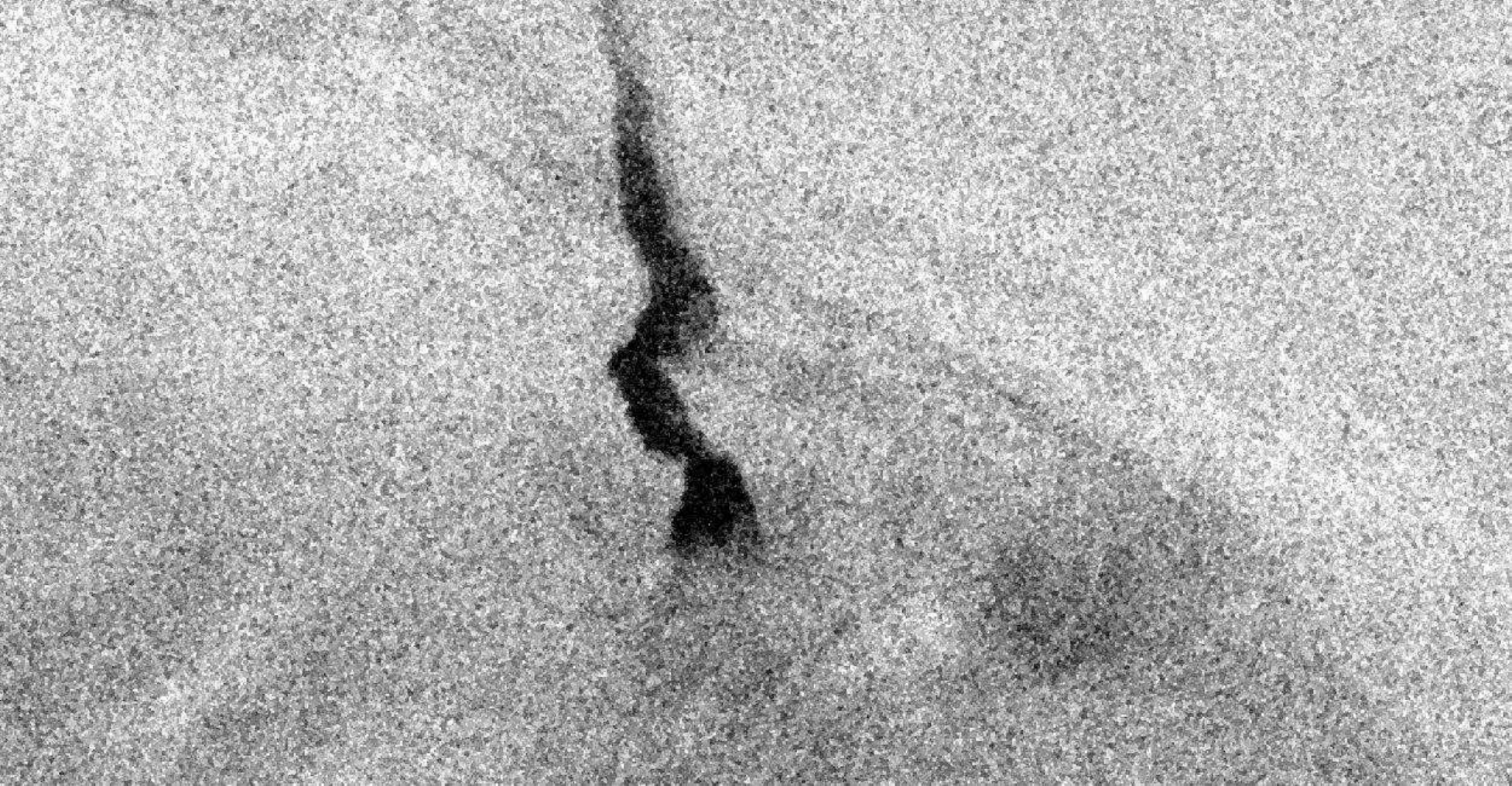}
        \caption{}
    \end{subfigure}
    \hfill
    \begin{subfigure}[b]{0.25\textwidth}
        \centering
        \includegraphics[width=\linewidth]{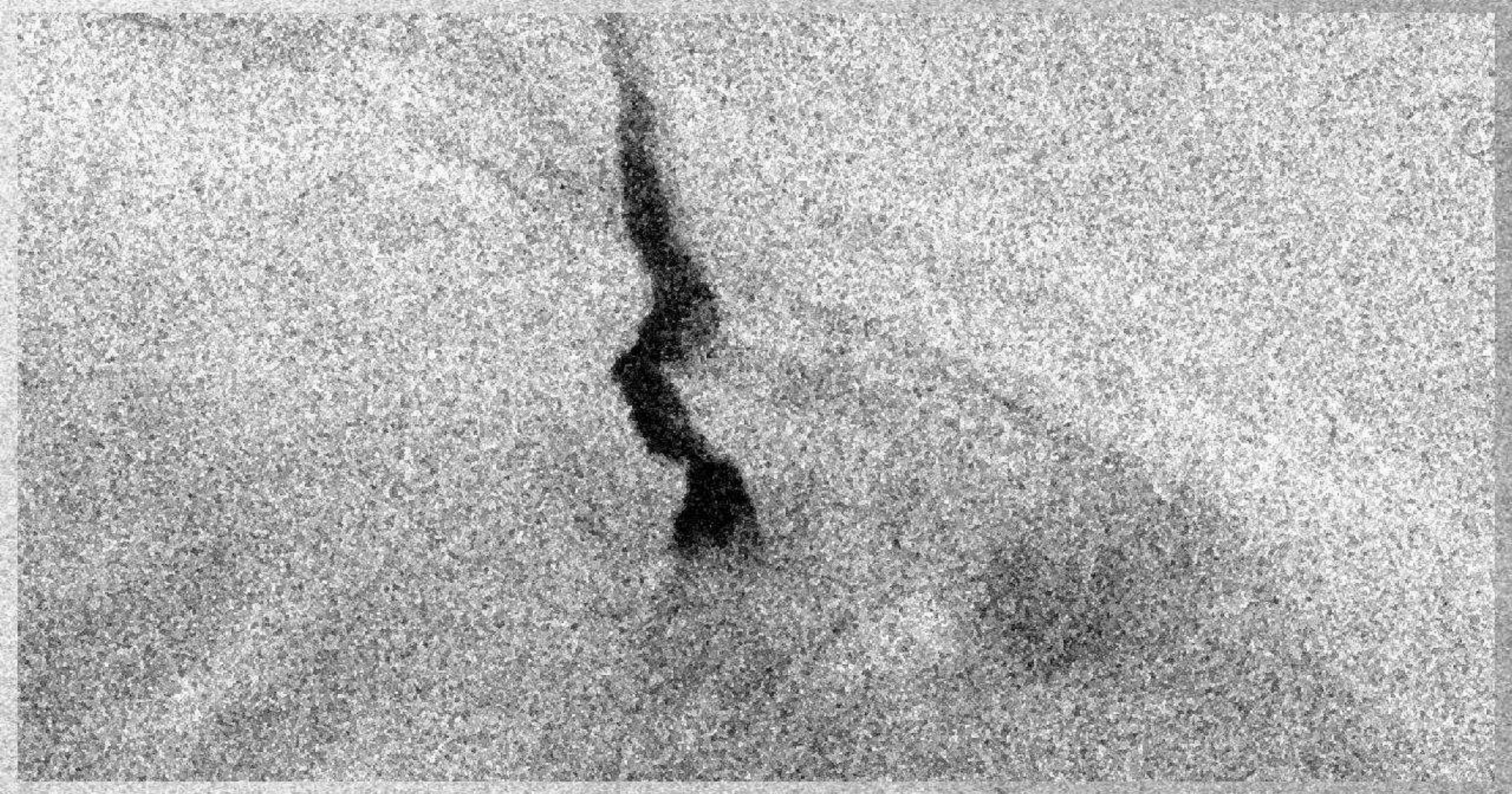}
        \caption{}
    \end{subfigure}
    \caption{(a) Example of an original image from the training set.  (b) A sample padded image from the training set.}
    \label{fig:padded_sample}
\end{figure}

In their research, Marios Krestenitis et al. used smaller image patches to train various models \cite{OilSpillDataset}. The smallest and largest image patch sizes used in their research were $320 \times 320$ and $336 \times 336$, respectively. In the work of this article, the original images, in $1250 \times 650$ dimensions, are padded with patches on all the $4$ sides of the image to produce resulting images of $1280 \times 672$. An original sample image from the training set and its corresponding padded image are shown in Figures \ref{fig:padded_sample}(a) and \ref{fig:padded_sample}(b), respectively. So, higher resolution images are used as input to the models, compared to those used by Marios Krestenitis et al. \cite{OilSpillDataset}. The patch padding is done in such a way as to select a patch of pixels with the sea surface. A random sample image from the training set is selected to make this patching task more straightforward. Data augmentation is applied to increase the size of the training dataset. Random horizontal and vertical flips are applied only on the training set for data augmentation. The images are normalized using the mean and standard deviation of the training set. Then, the normalized images are used as input for the various models used in this research.

\subsection{Models}
\subsubsection{Residual networks}
Increasing the depth of the convolutional neural networks (CNNs) beyond a certain point adversely affected the model's accuracy due to the vanishing and exploding gradient problems. To solve this, Residual Network (ResNet) \cite{res_net} was introduced by Kaiming He et al. The ResNet consists of residual blocks that can learn identity mappings in case of deeper networks where gradients can vanish. When stacked, these residual blocks help mitigate the vanishing gradient problem by learning an identity function.

Let us consider input $x$, and the desired mapping from input to output is denoted by $H(x)$. The residual denoted by $F(x)$ between the output $H(x)$ and input $x$ can be computed using the equation (\ref{eq:residue})

\begin{equation} 
    \label{eq:residue}
	F(x) = H(x) - x
\end{equation}

Hence, instead of using the original mapping, it can be recast into residual mapping using the equation (\ref{eq:residue_recast})

\begin{equation} 
    \label{eq:residue_recast}
    H(x) = F(x) + x
\end{equation}

To perform the above mapping, a constraint is that $F(x)$ and $x$ should be of the exact dimensions. But if that is not the case, then one can use a projection vector $W_s$ to match the dimensions. This is shown in equation (\ref{eq:residue_projection}).

\begin{equation} 
    \label{eq:residue_projection}
    y = F(x,\{W_i\}) + W_s x
\end{equation}

Kaiming He et al. \cite{res_net} showed that it was easier to learn the residual mapping than the original mapping. 

In this article, various pre-trained variants of ResNet models ---  ResNet-18, ResNet-34, ResNet-50, and ResNet-101 encoders are used, i.e., without the final classification layer that was used for object recognition on the ImageNet dataset \cite{imagenet_old,imagenet_new,res_net}.

\subsubsection{EfficientNet}
Neural Architecture Search (NAS) was proposed to optimize network architecture for image classification by Zoph et al. \cite{nas_image}. The training bottlenecks of EfficientNet were addressed, and EfficientNetV2 was proposed by Mingxing Tan et al. \cite{efficient_net_v1,efficient_net_v2}. The main change was replacing the MBConv block of EfficientNet with a new block proposed by Mingxing Tan et al., the FusedMBConv block, in the early layers. In the MBConv block, a $3 \times 3$ depthwise convolution layer was followed by a $1 \times 1$ normal convolution layer. This was replaced in the FusedMBConv block with a single fused $3 \times 3$ convolution layer. The other changes in EfficientNetV2 were to use $3 \times 3$ kernel sizes instead of larger kernel sizes but with more layers and a smaller expansion ratio for MBConv blocks since smaller expansion ratios and removal of the last stride-1 stage, which were optimizations towards a reduction of the memory access overhead \cite{efficient_net_v2}.

\subsubsection{DeepLabV3 \& DeepLabV3+}
The output of the encoder was provided as the input to the decoder, i.e., one of the decoders considered in this research --- DeepLabV3 and DeepLabV3+. 

In the DeepLabV3 decoder, Atrous Spatial Pyramid Pooling (ASPP) block was used \cite{deeplab_v3,deeplab}. This was a modification of earlier versions of the DeepLab decoder where atrous convolution layers were used instead of standard convolution layers. A dilation rate is used in atrous or dilated convolution, which uses a larger view of pixels when the kernel is applied to the image. In this ASPP block, there were four atrous convolution layers. ASPP has an average pooling layer applied on the feature maps from the encoder block to provide global context information. The outputs of all these five layers of the ASPP block were concatenated, and bilinear upsampling was applied to produce feature maps with the exact dimensions of the input image dimensions.

There were some minor changes to the DeepLabV3+ decoder when compared with that of DeepLabV2 \cite{deeplab_v3,deeplabv3+}. The outputs of all five layers of the ASPP block were concatenated, and bilinear upsampling by a factor of 4 was applied. To the corresponding features from the encoder block, a $1 \times 1$ convolution was used to balance the importance between the backbone's low-level features and the encoder block's compressed semantic features. The resulting features were concatenated with upsampled features followed by $3 \time 3$ convolution layers to refine the concatenated features. This connection from the encoder block and concatenation was a minor change in the DeepLabV3+ decoder. The resulting feature maps were upsampled using bilinear upsampling to produce feature maps with the exact dimensions of the input image dimensions.

\subsection{Training Models}\label{sec:training}
The following encoders are used in this research --- ResNet-18, ResNet-34, ResNet-50, ResNet-101, EfficientNetV2S, and EfficientNetV2M \cite{res_net,efficient_net_v2}. The pre-trained models trained on the ImageNet dataset are used for the encoder models using transfer learning \cite{transfer_learning,imagenet_old,imagenet_new}. The following decoders are used in this research --- DeepLabV3 and DeepLabV3+ \cite{deeplab_v3,deeplabv3+}. All the encoder-decoder models are trained end to end for 100 epochs, and the hyperparameters are obtained empirically.

Mean categorical cross-entropy loss is used since the dataset contained $5$ semantic classes. The categorical cross-entropy is given by Euqation \ref{eq:cate_cross_entropy} where $c_i$ denotes the encoded class and $p_i$ denotes the probability of the class as predicted by the model for every one of the $n$ classes in the dataset. The Stochastic Gradient Descent (SGD) optimizer is used with an initial learning rate of $1 \times 10^{-2}$, a momentum of $0.9$, and a weight decay of $1 \times 10^{-4}$. Liang-Chieh Chen et al. observed that the performance of the segmentation model was higher when SGD was combined with the Polynomial learning rate scheduler \cite{deeplab}. In this research, the SGD optimizer is combined with a polynomial learning rate scheduler, where the learning rate decays in a polynomial fashion. The Polynomial learning rate scheduler is given by \ref{eq:poly_lr_decay} where $lr_{0}$ is the initial learning rate, $e$ is the current epoch, $T_{e}$ is the total number of epochs, and $power$ controls the learning rate decay. Learning may be hampered once the learning rate at any epoch goes below a certain threshold and becomes closer to zero. To avoid this, a minimum learning rate is used as a threshold, and if the learning rate goes below the threshold, the threshold learning rate is used. For the Polynomial learning rate scheduler, the parameters --- $lr_{0}$ is set to $1 \times 10^{-2}$, $power$ is set to $0.9$ and the minimum learning rate is set to $1 \times 10^{-6}$.

\begin{equation}
    CE(c, p) = - \sum_{i=1}^{n} c_i \log(p_i)
    \label{eq:cate_cross_entropy}
\end{equation}

\begin{equation}
    lr = lr_{0} \times (1 - \frac{e}{T_{e}}) ^ {power}
    \label{eq:poly_lr_decay}
\end{equation}

A weight decay is used for regularization in the SGD optimizer, a dropout layer with a dropout rate of $10\%$, and data augmentation with horizontal and vertical flips is used.

Different batch sizes are used to train different models as they differ in the number of parameters that require different amounts of Graphical Processing Unit (GPU) memory. Table \ref{tab:batch_size} shows the batch size used to train different models. The Nvidia V100 GPU available on the high-performance computing cluster is used to train the models.

\begin{table}[t]
    \centering
    \caption{Batch sizes of different models used for training.}
    \scalebox{0.85}{
    \begin{tabular}{llll}
    \toprule
    \textbf{Encoder} & \textbf{Decoder} & \#\textbf{Params} & \textbf{Batch} \\
    & & \textbf{(millions)} & \textbf{size}\\
    \midrule
    ResNet-18 & DeepLabV3+ & 12.34 & 32\\
    ResNet-34 & DeepLabV3+ & 22.45 & 24\\
    ResNet-50 & DeepLabV3+ & 25.07 & 8\\
    ResNet-101 & DeepLabV3+ & 44.06 & 8\\
    EfficientNetV2S & DeepLabV3 & 21.42 & 8\\
    EfficientNetV2M & DeepLabV3 & 54.11 & 4\\
    \bottomrule
    \end{tabular}
    }
    \label{tab:batch_size}
\end{table}

\subsection{Transfer Learning}
Transfer learning is a machine learning approach that simplifies the training of deep neural networks from scratch \cite{transfer_learning}. Transfer learning allows us to use a previously developed machine learning model for a new but related task. This approach has gained significant popularity in the field of computer vision, primarily due to the impressive capability of CNNs to adapt learned low-level feature extraction for various tasks. 

\subsection{Evaluation Metrics}\label{sec:metric}
For evaluation of the performance of the models, Intersection over Union (IoU) is used \cite{OilSpillDataset}. The mean IoU (m-IoU) is the mean of the IoU of the different semantic classes in the dataset. The class IoU is the IoU computed for a semantic class individually.

\section{\uppercase{Results \& Discussion}\label{sec:results}}
\subsection{Quantitative Analysis}
A 5-fold cross-validation was performed to find the deviation of the performance of the models on different random validation splits. Table \ref{tab:validation_set_results} shows the performance metrics of the various models for a 5-fold validation on the randomized validation sets. For all the models, there is a significant deviation of m-IoU (greater than $1\%$) across the 5-fold validation. This is expected since the dataset was highly imbalanced and was split randomly into training and validation sets. Some splits would have higher m-IoU than others, depending on the division of the samples and their distribution of semantic classes. Hence, the m-IoU of the validation set depends on the split in cross-validation experiments.

\begin{table}[t]
    \centering
    \caption{Performance metrics of various models for a 5-fold validation on the randomized validation sets.}
    \scalebox{0.95}{
    \begin{tabular}{llll}
    \toprule
    \textbf{Encoder} & \textbf{Decoder} & \textbf{m-IoU} ($\%$) \\
    \midrule
    ResNet-18 & DeepLabV3+ & $67.345 \pm 1.407$ \\
    ResNet-34 & DeepLabV3+ & $67.522 \pm 3.898$ \\
    ResNet-50 & DeepLabV3+ & $67.578 \pm 3.631$ \\
    ResNet-101 & DeepLabV3+ & $68.152 \pm 2.470$ \\
    EfficientNetV2S & DeepLabV3 & $60.668 \pm 2.796$\\
    EfficientNetV2M & DeepLabV3 & $58.997 \pm 3.021$\\
    \bottomrule
    \end{tabular}
    }
    \label{tab:validation_set_results}
\end{table}

Table \ref{tab:test_set_results} shows the performance metrics of the best-performing model for each of the models on the test set with $110$ images. The best m-IoU on the test set was $64.868\%$ for the model with the ResNet-50 encoder and DeepLabV3+ decoder. This model's performance is slightly lower than the best-performing model from the previous research by Mario Krestenitis et al., which scored m-IoU of $65.06\%$ on the test set \cite{OilSpillDataset}. Table \ref{tab:batch_size} also shows the number of parameters in various models. If the model with EfficientNetV2M encoder and DeepLabV3 decoder with 54.11 million parameters and the model with ResNet-50 encoder and DeepLabV3+ decoder with just 25.07 million parameters are considered, their performances are $55.504\%$ and $64.868\%$ respectively. This shows that increasing the number of parameters in a model would not necessarily improve the model's performance for every task.

\begin{table}[!ht]
    \centering
    \caption{Performance metrics of the best-performing model for each model on the test set with $110$ images.}
    \scalebox{1}{
    \begin{tabular}{lll}
    \toprule
    \textbf{Encoder} & \textbf{Decoder} & \textbf{m-IoU} ($\%$) \\
    \midrule
    ResNet-18 & DeepLabV3+ & 59.647 \\
    ResNet-34 & DeepLabV3+ & 60.843 \\
    ResNet-50 & DeepLabV3+ & \textbf{64.868} \\
    ResNet-101 & DeepLabV3+ & 64.677 \\
    EfficientNetV2S & DeepLabV3 & 55.492 \\
    EfficientNetV2M & DeepLabV3 & 55.504 \\
    \bottomrule
    \end{tabular}
    }
    \label{tab:test_set_results}
\end{table}

Table \ref{tab:test_set_results_classwise} shows the classwise performance metrics of the best-performing models on the test set with $110$ images, i.e., the model with ResNet-50 encoder and DeepLabV3+ decoder from this research and best-performing model from Marios Krestenitis et al. research. Marios et al. best model scored a class IoU of $53.38\%$ and $55.40\%$ for the ``oil spill" and ``oil spill look-alike" classes, respectively \cite{OilSpillDataset}. On the other hand, the best-performing model from this research scored a class IoU of $61.549\%$ and $40.773\%$ for the ``oil spill" and ``oil spill look-alike" classes, respectively. Although the model from this research scored lower for the ``oil spill look-alike" class, it still scored higher for the ``oil spill" class, which is of more interest. Another observation is that the performance of detection of ``ship" is more remarkable for the best-performing model from this research when compared with that of the best-performing model from Marios Krestenitis et al., with class IoU of $33.378\%$ and $27.63\%$ respectively. For the remaining classes, i.e., the ``sea surface" and ``land", the performances of the best-performing models from this research and Marios Krestenitis et al. are comparable.

\begin{table}[!ht]
    \centering
    \caption{Classwise performance metrics, i.e., the class IoU of the best-performing model on the test set with $110$ images.}
    \scalebox{0.95}{
    \begin{tabular}{lll}
    \toprule
    & \textbf{Class IoU} ($\%$) & \\
    \midrule
    & \textbf{Our best} & \textbf{Best model of}\\
    & \textbf{model} & \textbf{Marios et al.}\\
    \midrule
    \textbf{Semantic Class} & & \\
    \midrule
    Sea surface & 96.422 & \textbf{96.43}\\
    Oil spill & \textbf{61.549} & 53.38\\
    Oil spill look-alike & 40.773 & \textbf{55.40}\\
    Ship & \textbf{33.378} & 27.63\\
    Land & 92.218 & \textbf{92.44}\\
    \bottomrule
    mean & 64.868 & \textbf{65.06}\\
    \bottomrule
    \end{tabular}
    }
    \label{tab:test_set_results_classwise}
\end{table}

\begin{figure}[!ht]
    \centering
    \begin{subfigure}[b]{0.35\textwidth}
        \centering
        \includegraphics[height=1in, width=\linewidth]{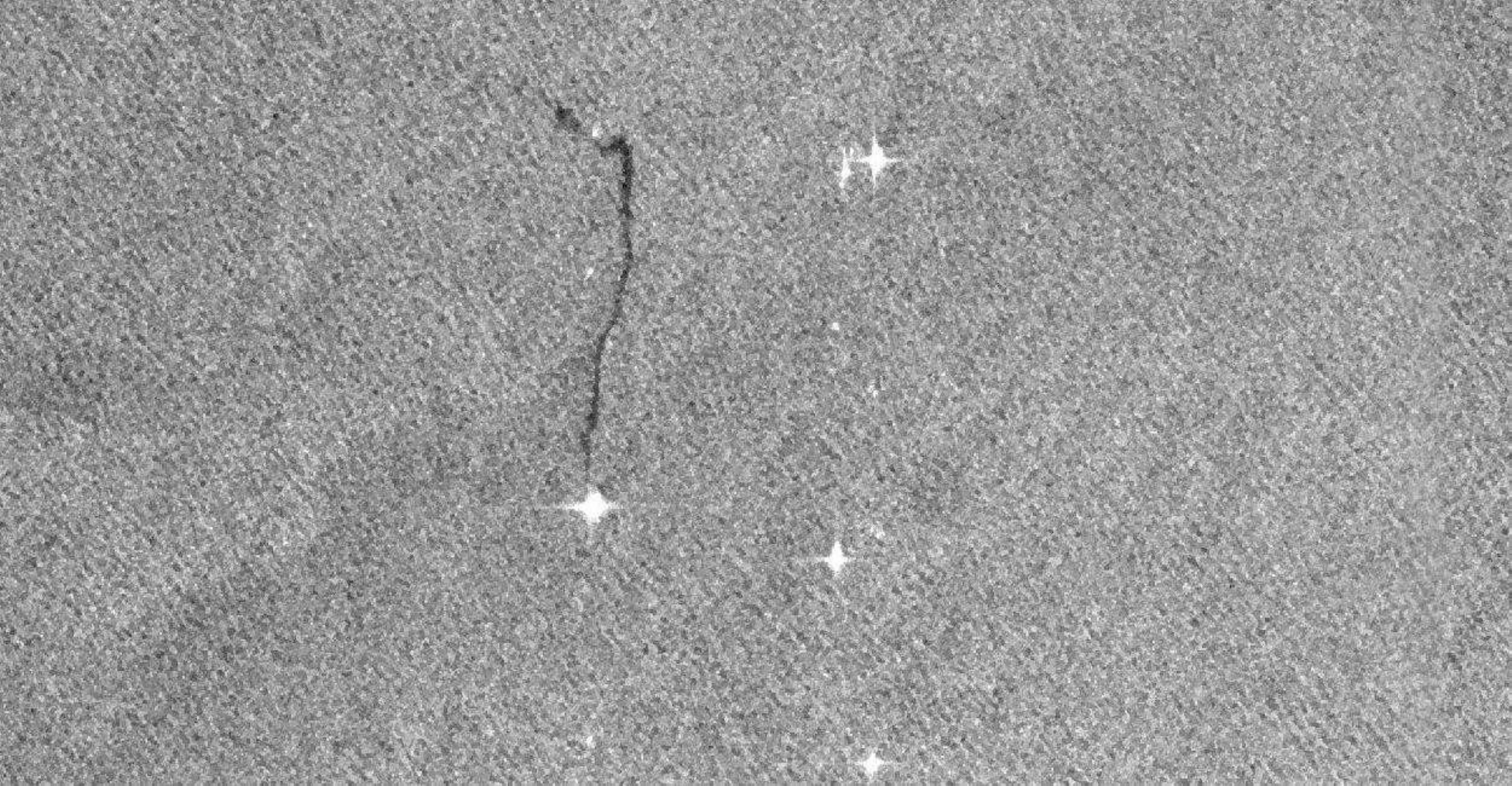}
        \caption{}
    \end{subfigure}
    \hfill
    \begin{subfigure}[b]{0.35\textwidth}
        \centering
        \includegraphics[height=1in, width=\linewidth]{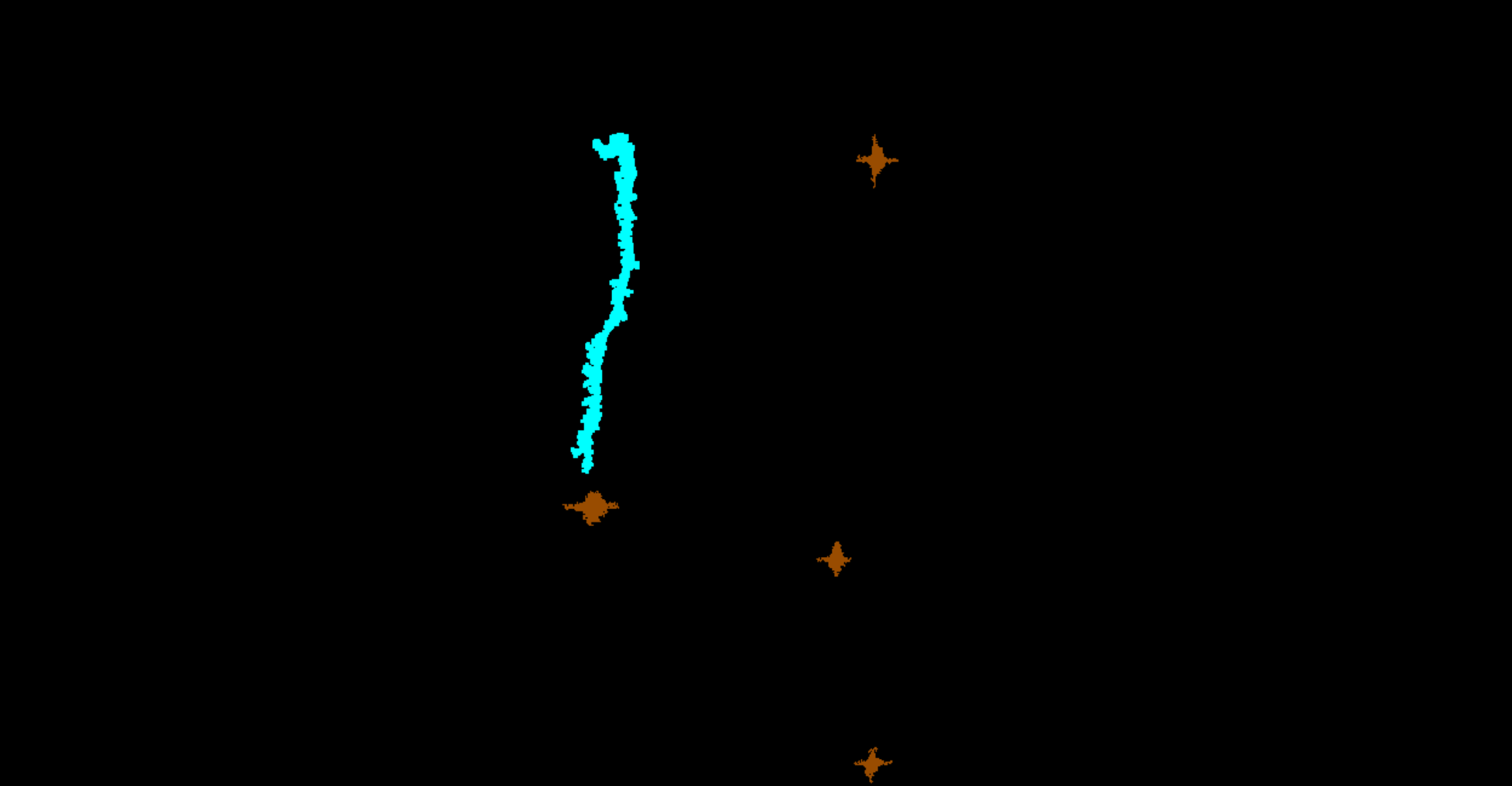}
        \caption{}
    \end{subfigure}
    \hfill
    \begin{subfigure}[b]{0.35\textwidth}
        \centering
        \includegraphics[height=1in, width=\linewidth]{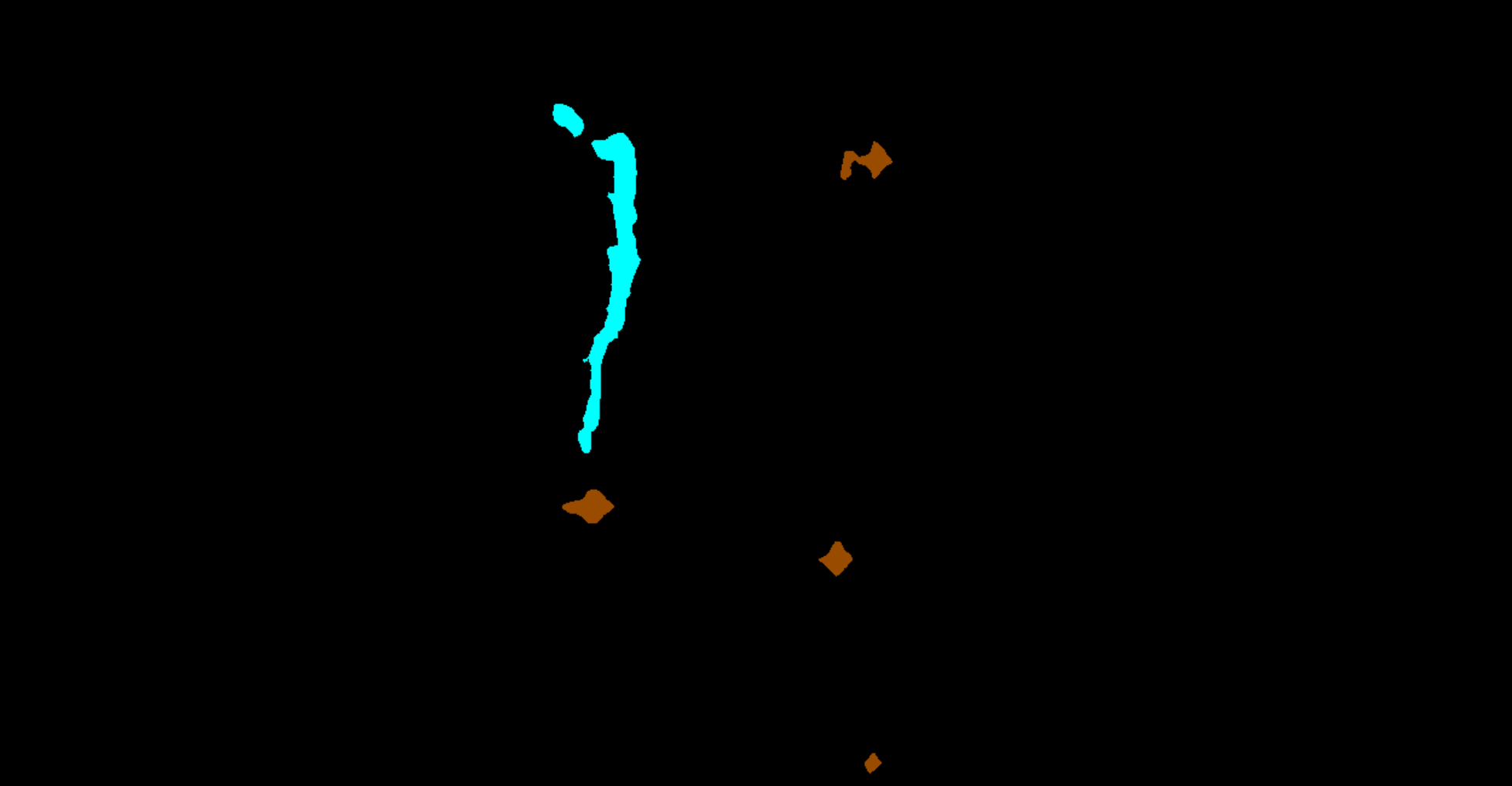}
        \caption{}
    \end{subfigure}
    \caption{(a) Sample test image  (b) Groundtruth  (c) Predicted mask.}
    \label{fig:test_sample_1}
\end{figure}

\begin{figure}[!ht]
    \centering
    \begin{subfigure}[b]{0.35\textwidth}
        \centering
        \includegraphics[height=1in, width=\linewidth]{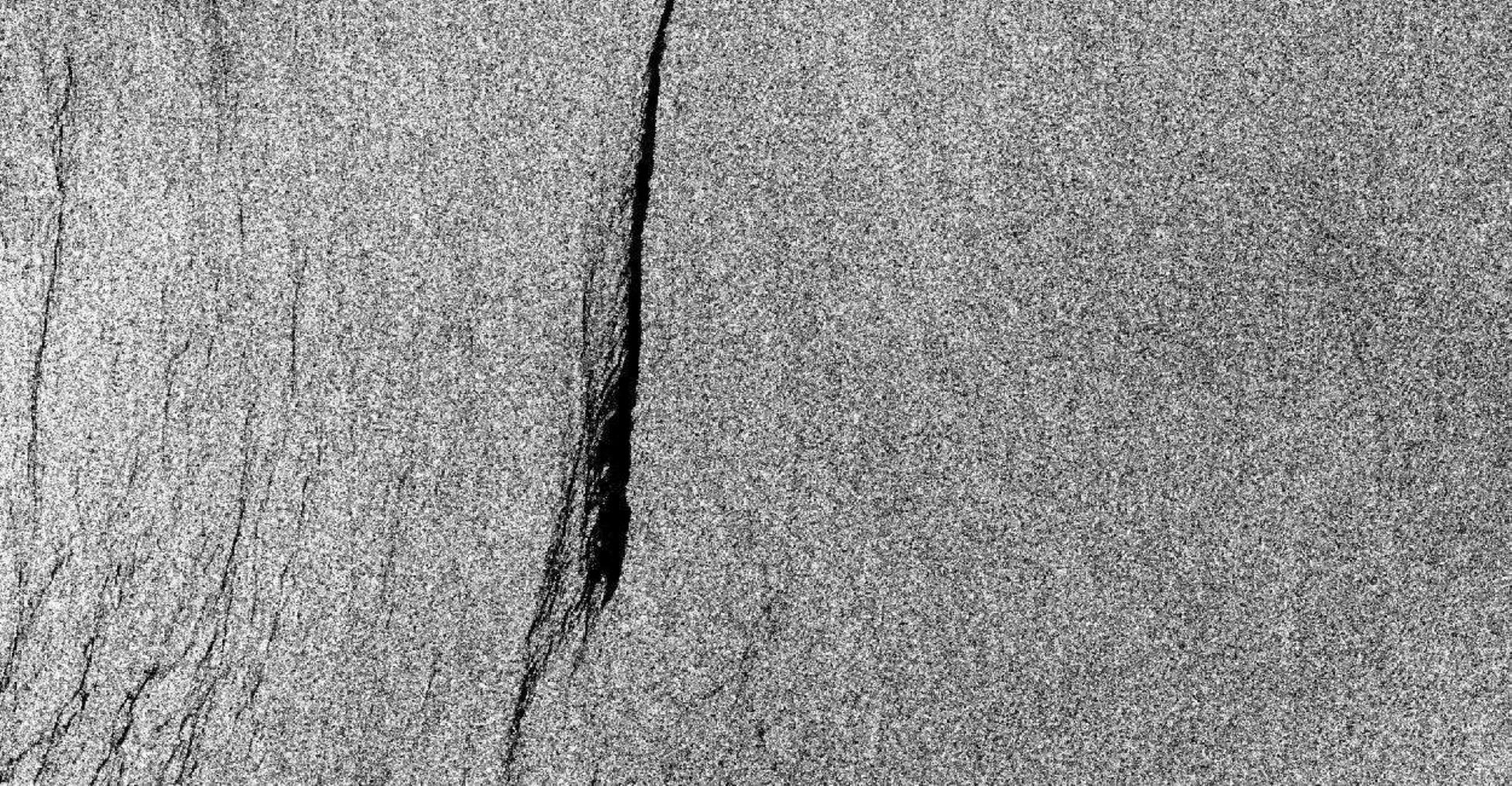}
        \caption{}
    \end{subfigure}
    \hfill
    \begin{subfigure}[b]{0.35\textwidth}
        \centering
        \includegraphics[height=1in, width=\linewidth]{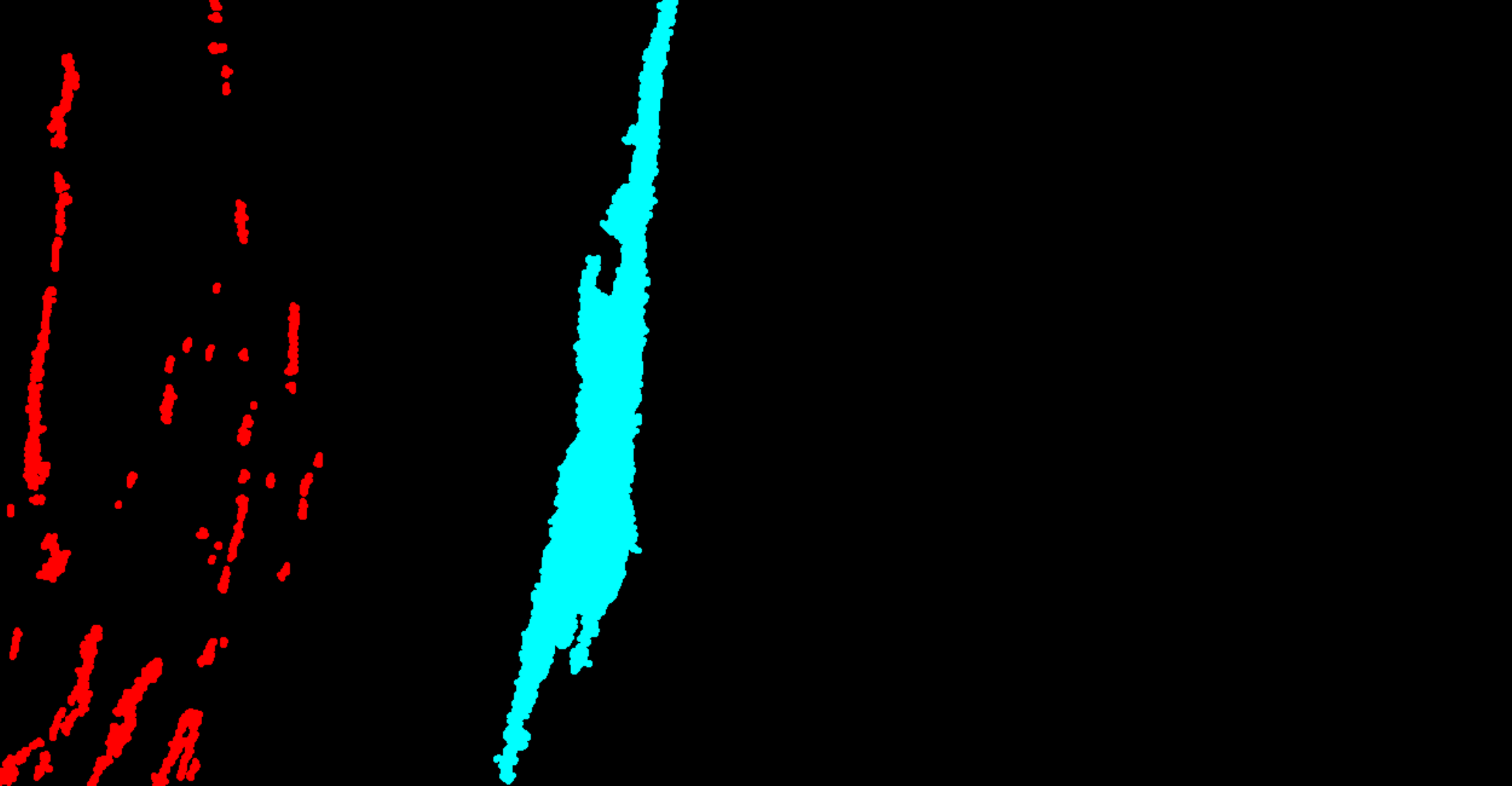}
        \caption{}
    \end{subfigure}
    \hfill
    \begin{subfigure}[b]{0.35\textwidth}
        \centering
        \includegraphics[height=1in, width=\linewidth]{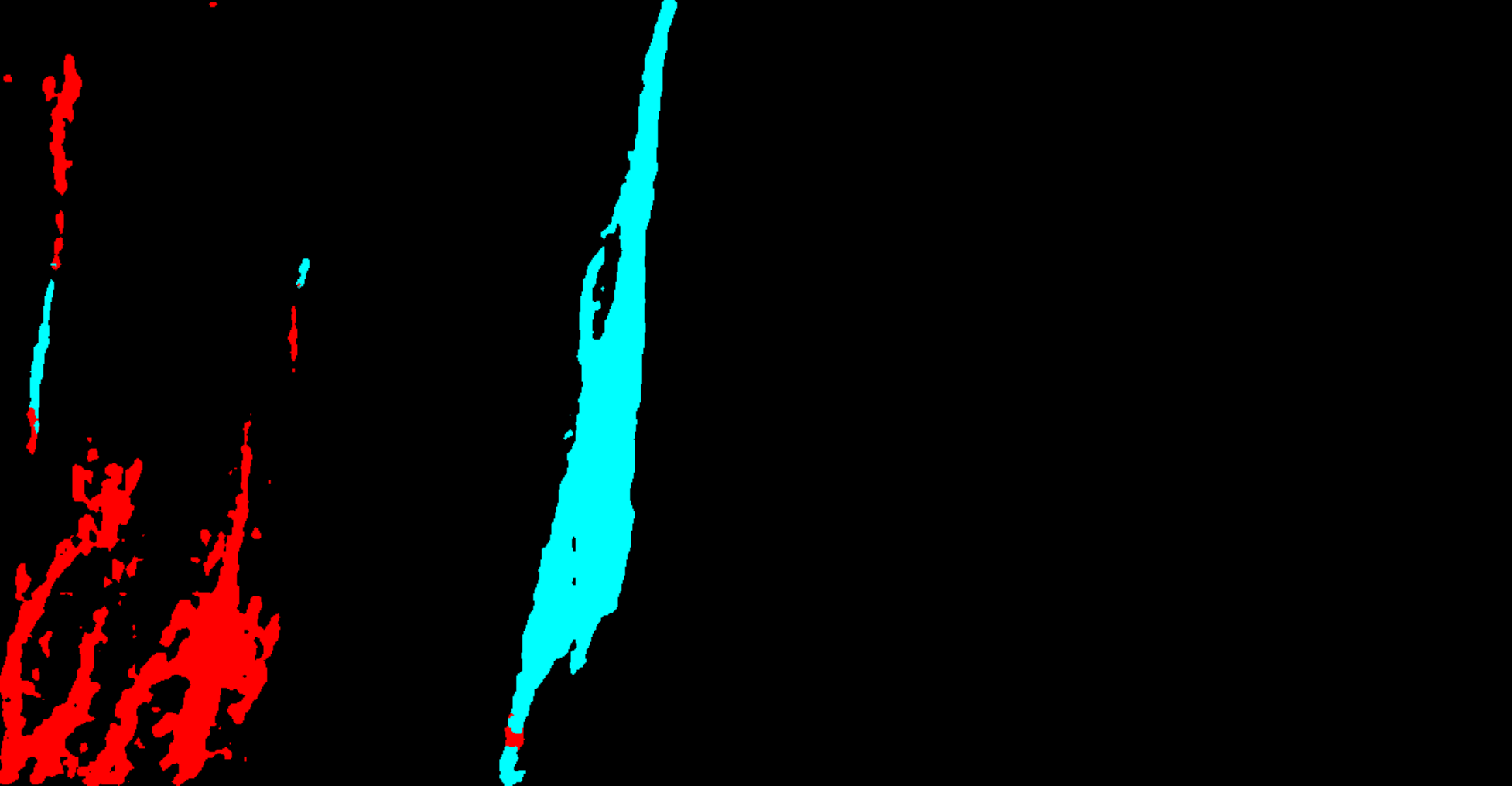}
        \caption{}
    \end{subfigure}
    \caption{(a) Sample test image  (b) Groundtruth  (c) Predicted mask.}
    \label{fig:test_sample_2}
\end{figure}

\begin{figure}[!ht]
    \centering
    \begin{subfigure}[b]{0.35\textwidth}
        \centering
        \includegraphics[height=0.9in, width=\linewidth]{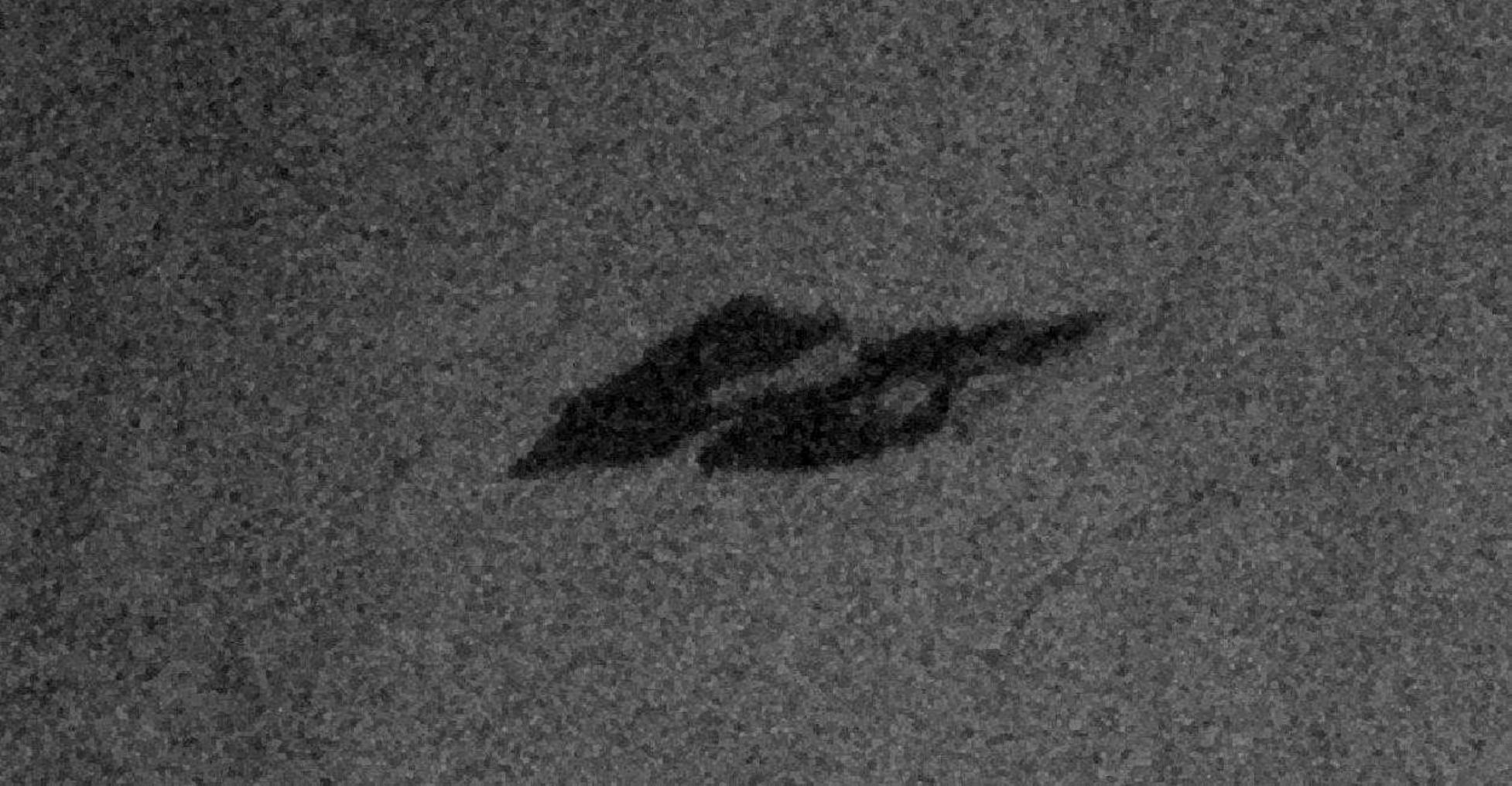}
        \caption{}
    \end{subfigure}
    \hfill
    \begin{subfigure}[b]{0.35\textwidth}
        \centering
        \includegraphics[height=0.9in, width=\linewidth]{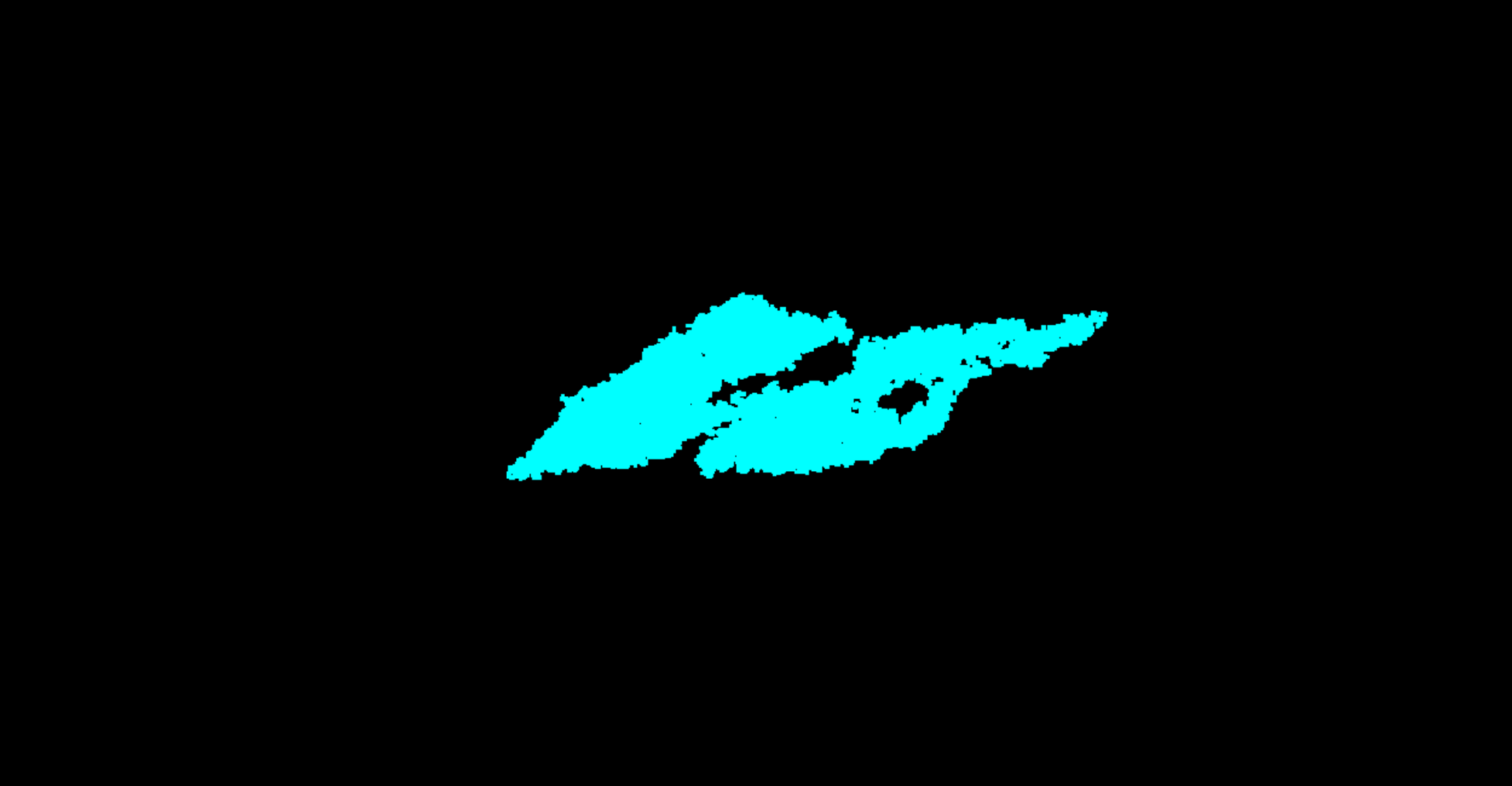}
        \caption{}
    \end{subfigure}
    \hfill
    \begin{subfigure}[b]{0.35\textwidth}
        \centering
        \includegraphics[height=0.9in, width=\linewidth]{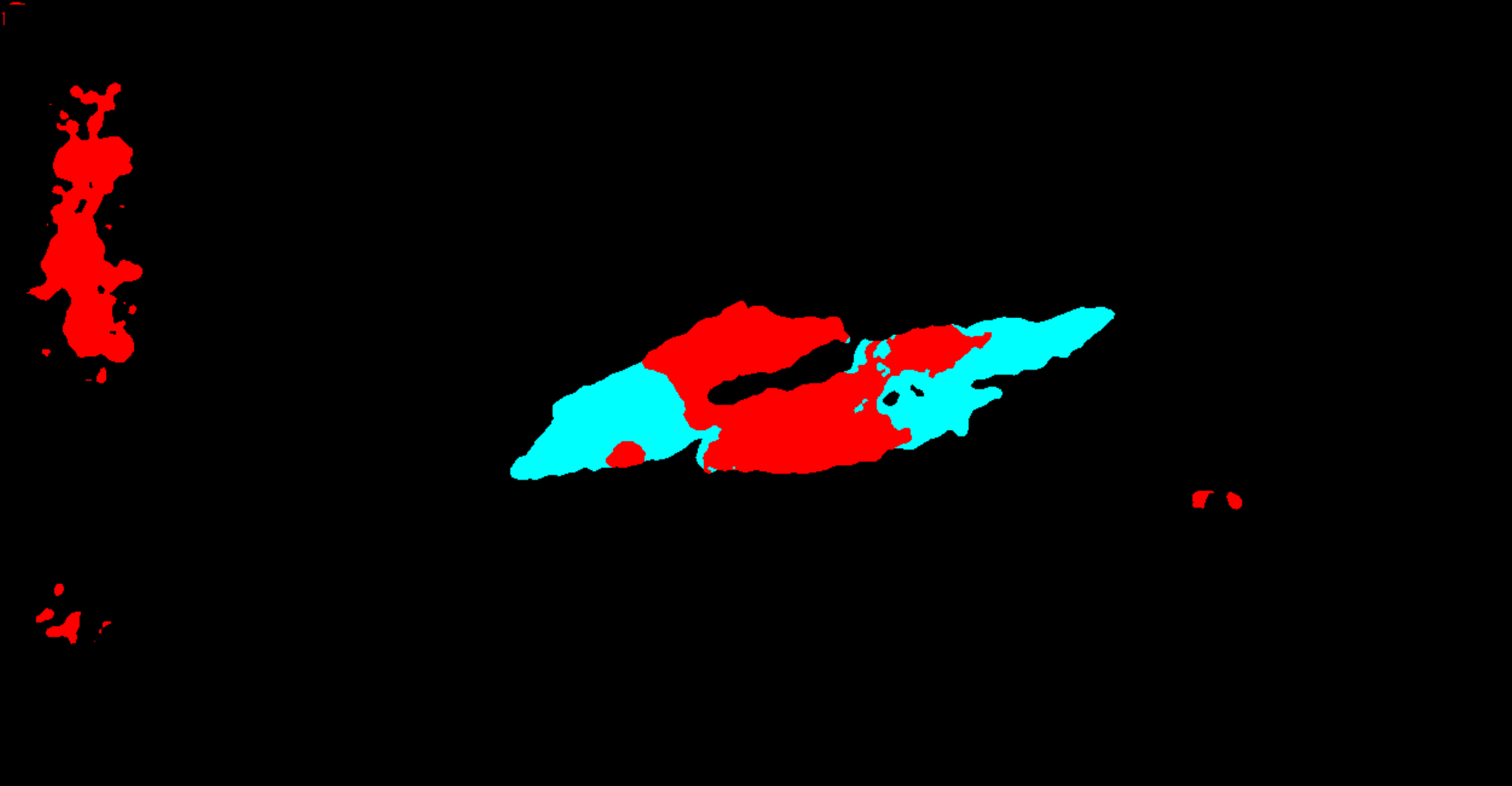}
        \caption{}
    \end{subfigure}
    \caption{(a) Sample test image  (b) Groundtruth  (c) Predicted mask.}
    \label{fig:test_sample_3}
\end{figure}



\subsection{Qualitative Analysis}
From the test sample shown in Figure \ref{fig:test_sample_1}(a), its ground-truth mask shown in Figure \ref{fig:test_sample_1}(b) and its prediction mask shown in Figure \ref{fig:test_sample_1}(c), it can be observed that relatively smaller areas of ``oil spills" (in Cyan) along with ``ships" (in Brown) are detected by the model with reasonable accuracy, for this sample. From Figures \ref{fig:test_sample_2}(a), \ref{fig:test_sample_2}(b), and \ref{fig:test_sample_2}(c), it can be observed that relatively more significant areas of ``oil spills" are detected by the model with reasonable accuracy. In this test sample, the model confuses the classes "oil spill" and "oil spill look-alike," resulting in portions of the regions being predicted for both classes. However, from Figures \ref{fig:test_sample_3}(a), \ref{fig:test_sample_3}(b), and \ref{fig:test_sample_3}(c), it can be observed that relatively more significant areas of ``oil spills" are \textbf{not} detected by the model with reasonable accuracy. The model showcases a range of detections, including highly accurate results as well as some less reliable ones.


\subsection{Learning Curves}
\begin{figure}[!ht]
    \centering
    \begin{subfigure}[b]{0.35\textwidth}
        \centering
        \includegraphics[width=\linewidth]{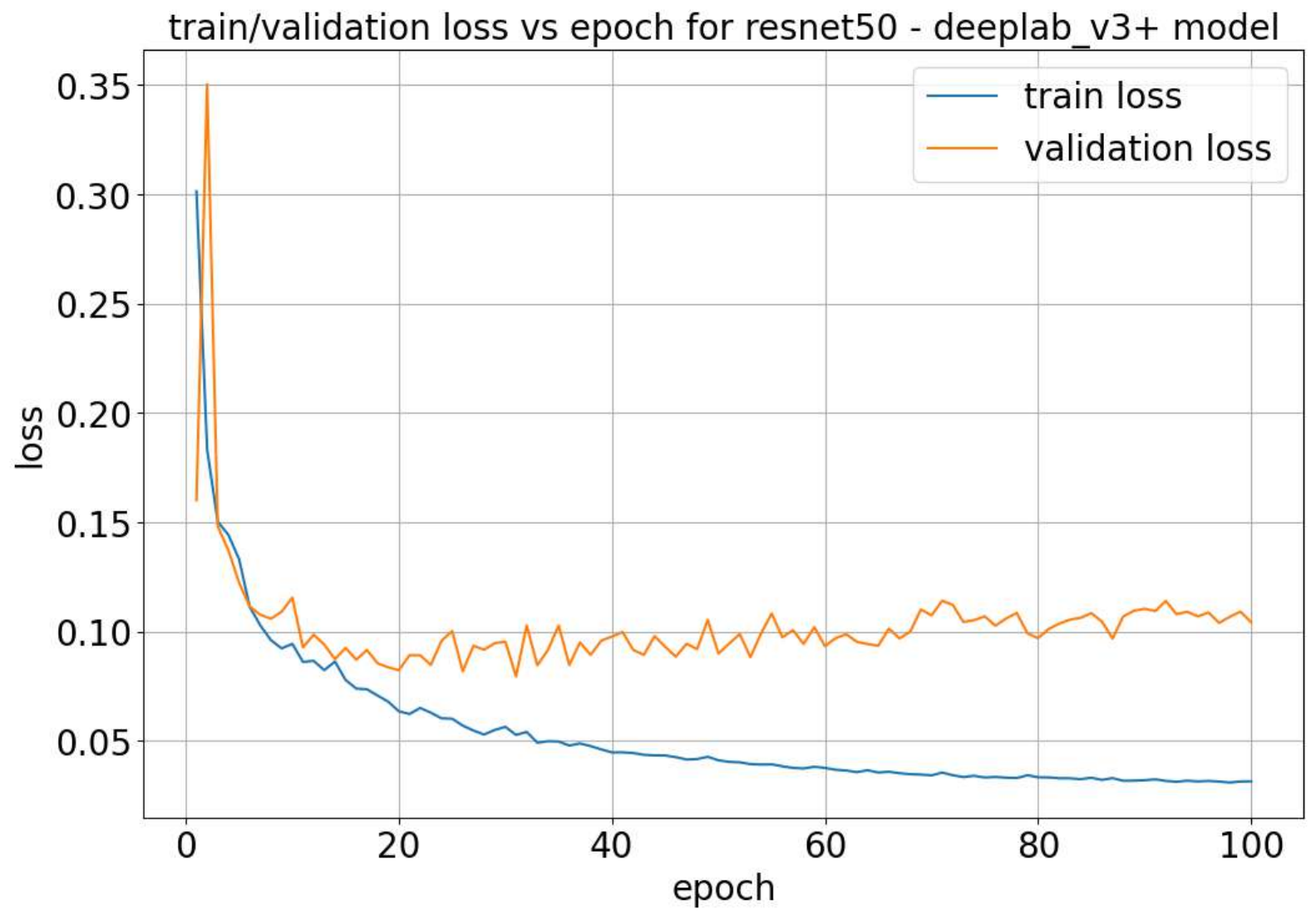}
        \caption{}
    \end{subfigure}
    \hfill
    \begin{subfigure}[b]{0.35\textwidth}
        \centering
        \includegraphics[width=\linewidth]{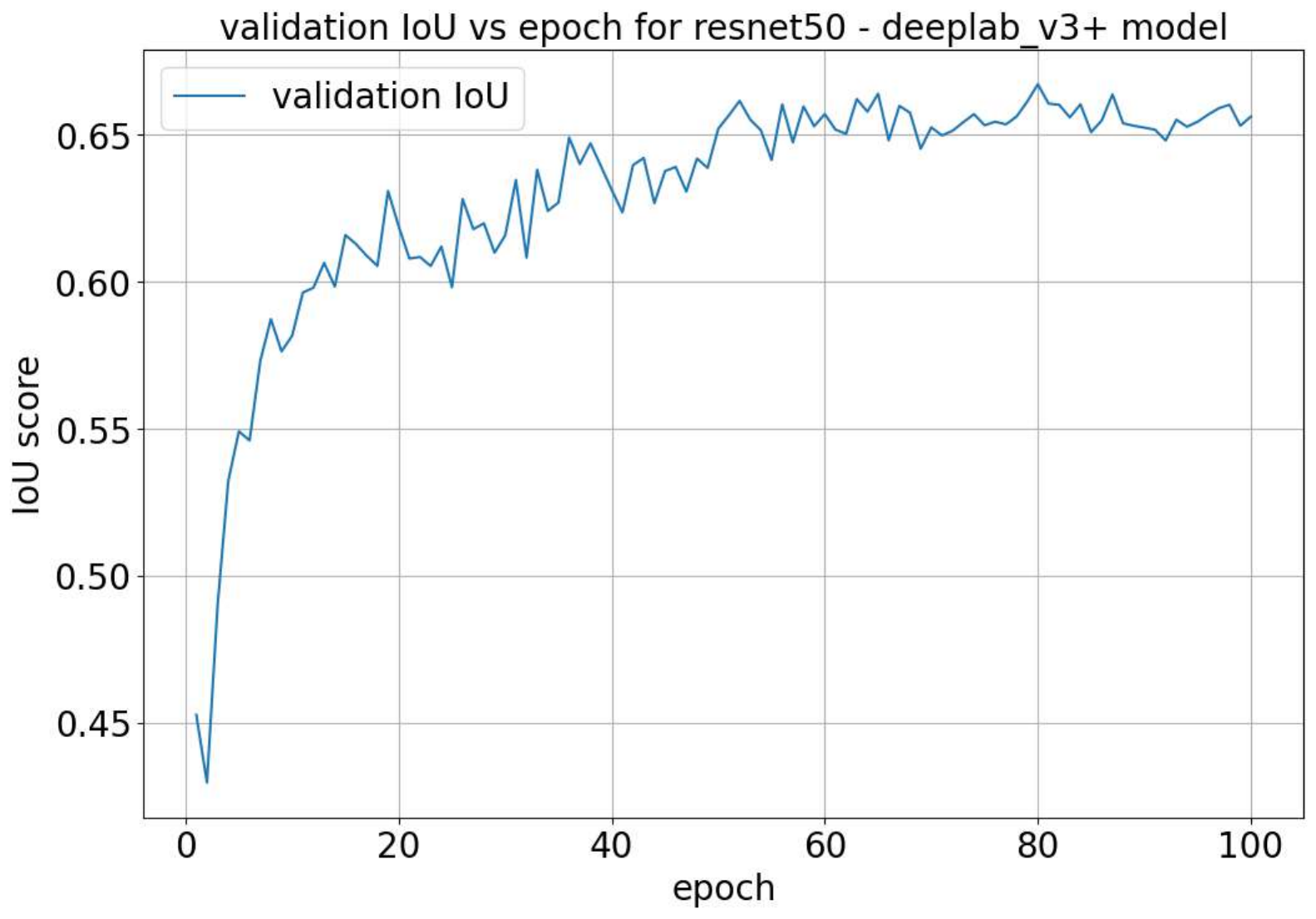}
        \caption{}
    \end{subfigure}
    \caption{(a) Plot of training and validation losses for the model with ResNet-50 encoder and DeepLabV3+ decoder.  (b) Plot of validation m-IoU for the model with ResNet-50 encoder and DeepLabV3+ decoder.}
    \label{fig:resnet_50_losses_iou}
\end{figure}

Figure \ref{fig:resnet_50_losses_iou}(a) shows the plot of losses for training and validation sets vs. epoch for the model with ResNet-50 encoder and DeepLabV3+ decoder, i.e., for the best-performing models from our experiments. \ref{fig:resnet_50_losses_iou}(b) shows the plot of m-IoU for validation set vs. epoch for the model with ResNet-50 encoder and DeepLabV3+ decoder. The two learning curve figures show that the validation metrics converge, i.e., they do not deviate much after considerable learning. However, some overfitting occurs as the gap between the training and validation losses widens with an increase in the number of epochs.

\section{\uppercase{Conclusions}\label{sec:conclusions}}
In the earlier research on the same topic and dataset, high-dimensional images were divided into smaller patches that served as input for the models. The best-performing model achieved a mean Intersection over Union (m-IoU) of $65.06\%$ and a class IoU of $53.38\%$. In contrast, this article utilized high-dimensional images as direct input to the model without partitioning them into patches. The best-performing model in this study, which employed a ResNet-50 encoder and a DeepLabV3+ decoder, achieved an m-IoU of $64.868\%$ and a class IoU of $61.549\%$ for the "oil spill" class. This represents a new benchmark result, indicating that using high-dimensional images can enhance the performance of "oil spill" detection.

However, the dataset also contains samples from the "oil spill look-alike" class, which can create confusion for the trained models, making it challenging to distinguish between the "oil spill" and "oil spill look-alike" classes. There is potential for further experimentation with various encoders and decoders to achieve better results. Additionally, the current encoders and decoders used in this research could be improved by incorporating visual self-attention modules. These enhancements can be explored in future research following the analysis presented in this article.

 \bibliographystyle{apalike} 

\section*{\uppercase{Acknowledgements}\label{sec:acknowledgements}}
We thank the Center for Information Technology of the University of Groningen for their support and for providing access to the Peregrine high-performance computing cluster. We also thank Marios Krestenitis and Konstantinos Ioannidis for creating and providing the dataset.

\end{document}